\documentclass[letterpaper]{article} 
\usepackage{aaai23}  
\usepackage{times}  
\usepackage{helvet}  
\usepackage{courier}  
\usepackage[hyphens]{url}  
\usepackage{graphicx} 
\usepackage{natbib}  
\usepackage{caption} 
\usepackage{algorithm}
\usepackage{algorithmic}

\usepackage{newfloat}
\usepackage{listings}
\DeclareCaptionStyle{ruled}{labelfont=normalfont,labelsep=colon,strut=off} 
\floatstyle{ruled}
\newfloat{listing}{tb}{lst}{}
\floatname{listing}{Listing}
\nocopyright

\title{Facial Affect Recognition based on Transformer Encoder and Audiovisual Fusion for the ABAW5 Challenge}

\author{
    Ziyang Zhang\footnotemark[1],
    Liuwei An\footnotemark[1],
    Zishun Cui,
    Ao Xu,
    Tengteng Dong,
    Yueqi Jiang, 
    Jingyi Shi,
    Xin Liu \\
    Xiao Sun\footnotemark[2],
    Meng Wang\footnotemark[2]
}

\affiliations{
    Hefei University of Technology, Hefei, China \\
    \{zzzzzy, anliuwei, 2022171285, 2022111070, 2022180026, 2017217795, 2019213576, 2019218059\}@mail.hfut.edu.cn \\ 
    sunx@hfut.edu.cn\quad eric.mengwang@gmail.com
}

\usepackage{bibentry}

\begin{document}

\maketitle

\renewcommand{\thefootnote}{\fnsymbol{footnote}} 
\footnotetext[1]{These authors contributed equally.} 
\footnotetext[2]{Corresponding Author.} 
\renewcommand{\thefootnote}{\arabic{footnote}} 

\begin{abstract}
In this paper, we present our solutions for the 5th Workshop and Competition on Affective Behavior Analysis in-the-wild (ABAW),  which includes four sub-challenges of Valence-Arousal (VA) Estimation, Expression (Expr) Classification, Action Unit (AU) Detection and Emotional Reaction Intensity (ERI) Estimation. The 5th ABAW competition focuses on facial affect recognition utilizing different modalities and datasets. In our work, we extract powerful audio and visual features using a large number of sota models. These features are fused by Transformer Encoder and TEMMA. Besides, to avoid the possible impact of large dimensional differences between various features, we design an Affine Module to align different features to the same dimension. Extensive experiments demonstrate that the superiority of the proposed method. For the VA Estimation sub-challenge, our method obtains the mean Concordance Correlation Coefficient (CCC) of 0.6066. For the Expression Classification sub-challenge, the average F1 Score is 0.4055. For the AU Detection sub-challenge, the average F1 Score is 0.5296. For the Emotional Reaction Intensity Estimation sub-challenge, the average pearson’s correlations coefficient on the validation set is 0.3968. All of the results of four sub-challenges outperform the baseline with a large margin.
\end{abstract}

\section{Introduction}
\label{sec:intro}
Sentiment analysis is an important research field of pattern recognition, which introduces affective dimension into human-computer interaction. There are a number of related applications spread across a variety of fields, such as auxiliary treatment of mental illness, detection of fatigue driving, analysis of consumer attitudes, etc\cite{AlbertHaque2018MeasuringDS,GulbadanSikander2019DriverFD,VaradaKolhatkar2020TheSO}. 

Among the various affective factors, facial information is the most intuitive and realistic. In this paper, we focus on facial affect recognition and extensively present our solutions for the 5th Workshop and Competition on Affective Behavior Analysis in-the-wild (ABAW), which includes four sub-challenges of Valence-Arousal (VA) Estimation Challenge, Expression (Expr) Classification Challenge, Action Unit (AU) Detection Challenge and Emotional Reaction Intensity (ERI) Estimation Challenge. 

There are mainly two paradigms of emotions used in the emotion recognition, i.e. categorical and dimensional. Categorical emotions are based on the six basic emotions proposed by Ekman (happiness, sadness, fear, anger, disgust, and surprise) \cite{PaulEkman1992AnAF}. Dimensional approaches mainly use two dimensions : valence and arousal which are based on Russell's circumplex model of emotions\cite{JamesARussell1980ACM}. Valence refers to how negative to positive the person feels and arousal refers to how sleepy to active a person feels. 

For Valence-Arousal (VA) Estimation Challenge, the level of valence and arousal in a time-continuous manner needs to be predicted from audio-visual recordings. It is necessary to predict the values of valence and arousal in a continuous way using the database which consists of around 3M frames annotated in terms of valence and arousal.

For Expression (Expr) Classification Challenge, we need to train a model to predict 6 basic expressions, the neutral state and a category 'other' using the database which consists of around 2.7M frames that are annotated in terms of the 6 basic expressions (i.e. anger, disgust, fear, happiness, sadness, surprise),plus the neutral state,plus a category 'other' that denotes expressions/affective states other than the 6 basic ones.

For Action Unit (AU) Detection Challenge, the facial action coding system(FACS)\cite{PaulEkman2019FacialAC} defines a group of action units (AU) from the perspective of facial anatomy, which is used to accurately describe the changes of facial expression. Each facial action unit describes the apparent changes caused by a group of facial muscle movements, which can express any facial expression. AU detection is a multi-label classification problem. Its challenges lie in insufficient labeling data, head posture interference, individual differences, and the imbalance of different AU categories.We need to train a model to predict whether AUs are active using the database which consists of around 2.7M frames that are annotated in terms of 12 action units. 

For Emotional Reaction Intensity (ERI) Estimation Challenge, compared with  facial expression classification, Emotional Reaction Intensity (ERI) Estimation is more challenging. For people of different ages, genders, and cultural backgrounds, their emotional experience often has great differences, which makes the Emotional Interaction Intensity (ERI) Estimation more complex, and requires muti-modal information to make complex emotional recognition. We need to train a muti-model to predict 7 emotional experiences using the Hume-Reaction dataset which consists of subjects reacting to a wide range of various emotional video-based stimuli including Adoration, Amusement, Anxiety, Disgust, Empathic Pain, Fear, and Surprise.

The main contribution of the proposed method can be summarized as:
\begin{enumerate}
    \item We extract potentially useful information from two modalities, audio and visual, respectively. For the audio modality, we extract audio features using the correlation toolkit. For the visual modality, we use many sota models to extract the corresponding emotional features or facial features.
    \item To mitigate the possible impact of large dimensional differences between features, we design an Affine Module to align different features to the same dimension and add positional encoding to convey the contextual relationship of the sequence to the model.
    \item We explored all four challenges of ABAW5, and although for the encoder part we used all common Transformer encoder, we achieved good performance by adding Affine Module, already with a reasonable combination of features.
\end{enumerate}

\section{Related Work}
\label{sec:relat}

\subsection{Multimodal Features}
The utilization of multimodal features, including visual, audio, and text features, has been extensively employed in previous ABAW competitions \cite{zafeiriou2017aff,kollias2019face,kollias2019expression,kollias2020analysing,kollias2021affect,kollias2021analysing,kollias2021distribution,kollias2022abaw,kollias2023abaw,kollias2023abaw2}. We can improve the performance in affective behavior analysis tasks by extracting and analyzing these multimodal features.

In the visual modality, the facial expression is an important aspect to understand and analyze emotions. In the Facial Action Coding System (FACS) \cite{ekman1978facial} proposed by Friesen and Ekman in 1978, the human face is represented by a set of specific facial muscle movements known as Action Units (AUs) and it has been widely applied in studies of facial expressions\cite{martinez2017automatic}. With the development of deep learning, it has been found that visual features based on convolutional\cite{poria2016convolutional} and transformer\cite{vaswani2017attention,han2022survey} networks can achieve better results.

In the context of affective computing, audio features, which typically include energy features, time-domain features, frequency-domain features, psychoacoustic features, and perceptual features, have been extensively utilized and shown to achieve promising performance in tasks such as expression classification and VA estimation\cite{zhang2018attention,stuhlsatz2011deep,lieskovska2021review}.These  features can be extracted through pyAudioAnalysis \cite{giannakopoulos2015pyaudioanalysis}, which is a Python library covering a wide range of audio analysis tasks. Similar to visual features, deep learning has also been widely used in acoustic feature extraction. 

The text modality has been increasingly explored to address emotion recognition tasks. To enhance the effectiveness of text modality, Word2Vec \cite{mikolov2013distributed} and GloVe \cite{pennington2014glove} have been proposed and demonstrated to achieve superior performance.In ABAW3 competition, Zhang et al.\cite{zhang2022transformer} utilized a word embedding extractor to extract text features, achieving promising results.

In the previous editions of the ABAW competition, many teams utilized multimodal features\cite{kollias2022abaw,jiang2022facial,jin2021multi,meng2022multi,zhang2022transformer,zhang2022continuous}.The model proposed by Meng et al.\cite{meng2022multi} leverages both audio and visual features, ultimately achieving first place in the VA track . To fully exploit the in-the-wild emotion information, Zhang et al.\cite{zhang2022transformer} utilizes the multimodal information from the images, audio and text and propose a unified multimodal framework for AU detection and expression recognition.The proposed framework achieved the highest scores on both tasks. These approaches convincingly demonstrate the effectiveness of multimodal features in affective behavior analysis tasks.

\subsection{Multimodal Structure}
In early studies, Zadeh et al.\cite{zadeh2016multimodal} and Pérez-Rosa et al.\cite{perez2013utterance} employed concatenated multimodal features to train Support Vector Machine (SVM) models, which were inadequate in effectively modeling the multimodal information. Recent research on multimodal emotion analysis has mainly used deep learning models to model both intra-modal and inter-modal information interactions. Truong et al.\cite{truong2019vistanet} develops a neural network model called Visual Aspect Attention Network (VistaNet), which takes visual information as the alignment source at the sentence level. This multimodal structure enables the model to pay more attention to these sentences when classifying emotions.Currently, the use of Transformer for multimodal learning has become mainstream in multimodal algorithms. In the field of image-text matching, ALBEF\cite{li2021align} is to some extent inspired by the CLIP\cite{radford2021learning} model, introducing the idea of multimodal contrastive learning into multimodal models, achieving the unity of multimodal contrastive learning and multimodal fusion learning.
In the previous ABAW competition, \cite{kim2022facial,tallec2022multi,wang2022multi,zhang2022transformer} utilizes transformer structures and achieves outstanding performance.

\subsection{Multimodal Fusion}
Multimodal research places great importance on fusion, the process of combining information extracted from multiple unimodal data sources into a unified and compact multimodal representation. Fusion methods are typically categorized according to the different stages in which fusion occurs, including early fusion, late fusion, hybrid fusion and so on\cite{zhang2020multimodal}.

During early fusion, the features are directly combined into general feature vectors for analysis by the model.Nagrani et al.\cite{nagrani2021attention}directly inputs a sequence of visual and auditory patches into the transformer. This early fusion model allows attention to flow freely between different spatial and temporal regions of the image, as well as across frequency and time in the audio spectrogram.

In the late fusion , each modality's features are independently analyzed, and their final outputs are then fused to acquire a better prediction.Zhang et al. \cite{zhang2022transformer} propose a unified late fusion framework for both Action Unit (AU) detection and expression recognition, leveraging multimodal information from images, audio, and text. The late fusion framework effectively incorporates prior multimodal knowledge, enabling effective emotion analysis from different perspectives and leading to the championship in both the AU and Expression tracks of ABAW3 competition.

As a fusion of both, Hybrid fusion combines the advantages of both early fusion and late fusion. Li et al. \cite{li2022hybrid} proposes a hybrid fusion method which leads them to the winner of MuSe-Reaction competition. In \cite{li2022hybrid}, audio features , facial expression features and  paragraph-level text embeddings are fused at the feature level and  then fed into the MMA module to extract complementary information from different modalities and calculate interactions between modalities.

\section{Feature Extraction}
\subsection{Audio Features}
\textbf{IS09:}The INTERSPEECH 2009(IS09) feature set was introduced at the INTERSPEECH 2009 Emotion Challenge\cite{schuller2009interspeech}, and it consists of 384 features that are derived from statistical functions applied to low-level descriptor contours. To extract these features, we utilized the openSMILE toolkit\cite{schuller2013interspeech}.

\textbf{VGGish:}VGGish\cite{hershey2017cnn} is a pre-trained neural network by Google for extracting speech-related features from audio signals. Its output is a 128-dimensional feature vector that can be used for speech-related tasks.

\textbf{eGeMAPS:}The extended Geneva Minimalistic Acoustic Parameter Set (eGeMAPS)\cite{zhao2018multi} is an extension of GeMAPS. The audio feature set in eGeMAPS is designed based on expert knowledge. eGeMAPS has only 88-dimensional features compared to traditional high-dimensional feature sets, but it shows higher robustness to speech emotion modeling problems.

\textbf{DeepSpectrum:}DeepSpectrum\cite{amiriparian2017snore} is a method of extracting deep spectrum features from audio file spectrograms using pre-trained convolutional neural networks (CNNs). These features are obtained by forwarding spectrograms through very deep task-independent pre-trained CNNs, and extracting the activations of fully connected layers as feature vectors. The dimension of the audio feature vectors is 1024.

\textbf{CNN14:}To obtain the high-level deep acoustic representations, a supervised model called PANNs\cite{gemmeke2017audio} is utilized, which has been pre-trained on the AudioSet dataset\cite{gemmeke2017audio}. PANNs comprises of multiple systems, and for this purpose, the CNN14 system that was trained using 16 kHz audio recordings is employed to generate a feature vector consisting of 2048 dimensions.

\subsection{Visual Features}
\textbf{EAC:}Erasing Attention Consistency (EAC)\cite{zhang2022learn} is a novel approach for addressing noisy samples during model training. The method leverages the flip semantic consistency of facial images to create an imbalanced framework, and randomly erases input images to prevent the model from overemphasizing specific features. By using flip attention consistency, EAC outperforms state-of-the-art methods for dealing with noisy labeled facial expression recognition (FER) datasets, and also generalizes well to other tasks with a large number of classes.

The EAC method, based on the ResNet50, achieves a high accuracy rate of 90.35\% on the RAF-DB\cite{li2017reliable} dataset, and the dimension of the resulting visual feature vector is 2048.

\textbf{ResNet-18:}ResNet\cite{he2016deep} is a deep learning architecture that addresses the vanishing gradient problem in deep neural networks by introducing residual connections. These connections create shortcuts across the network, allowing the input signal to bypass multiple layers and directly propagate to the deeper layers. This helps the network learn much deeper representations. In this study, we trained ResNet-18 using the AffectNet\cite{mollahosseini2017affectnet} dataset and obtained a 512-dimensional visual feature vector.

\textbf{POSTER:}The two-stream Pyramid crOss-fuSion TransformER network (POSTER)\cite{zheng2022poster} is proposed to address the challenges of facial expression recognition. It effectively integrates facial landmark and direct image features using a transformer-based cross-fusion paradigm and employs a pyramid structure to ensure scale invariance. Extensive experimental results demonstrate that POSTER outperforms SOTA methods on RAF-DB with 92.05\%, AffectNet (7 cls) with 67.31\%, and AffectNet (8 cls) with 63.34\%, respectively . The dimension of the visual feature vectors is 768.

\textbf{POSTER2:}The proposed POSTER2\cite{mao2023poster} aims to improve upon the complex architecture of POSTER, which achieves state-of-the-art performance in facial expression recognition (FER) by combining facial landmark and image features through a two-stream pyramid cross-fusion design. POSTER2 reduces computational cost by using a window-based cross-attention mechanism, removing the image-to-landmark branch in the two-stream design, and combining images with landmark's multi-scale features. Experimental results show that POSTER2 achieves state-of-the-art FER performance with minimum computational cost on several standard datasets. For example, POSTER2 achieves 92.21\% on RAF-DB, 67.49\% on AffectNet (7 cls), and 63.77\% on AffectNet (8 cls) using only 8.4G FLOPs and 43.7M Params. The same visual feature dimension as POSTER is 768.

\textbf{FAU:}Facial Action Units (FAU), originally introduced by Ekman and Friesen\cite{ekman1978facial}, are strongly associated with the expression of emotions . Therefore, the detection of FAU has become a popular and promising method for predicting affect-related targets. We use the OpenFace\cite{baltrusaitis2018openface} open source framework to extract FAU features and get a 17-dimensional feature vector.

\section{Method}
The 5th Competition on ABAW included a total of 4 challenges and we participated in all of them. Although the tasks for each challenge were not the same, our model for each challenge followed a basic framework. In detail, for the former three challenges, we refer to the design of the classical Transformer model\cite{AshishVaswani2017AttentionIA}, and for the fourth challenge, we adopt the Transformer Encoder with Multimodal Multi-Head Attention (TEMMA)\cite{HaifengChen2021TransformerEW} model.

\begin{figure*}[t]
\centering
\includegraphics[width=0.8\linewidth]{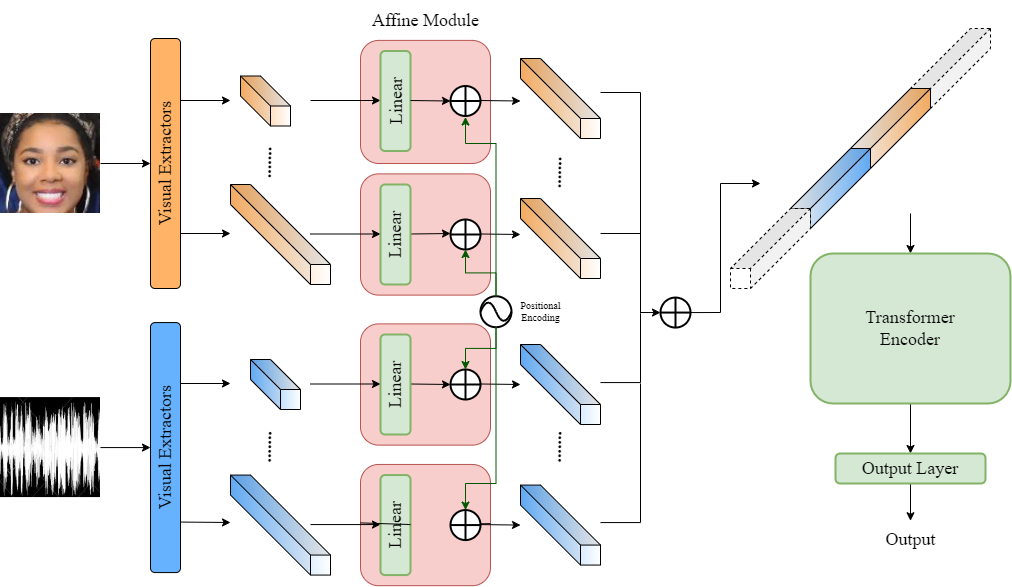}
\caption{The overall framework of our proposed method.}
\label{fig:model}
\end{figure*}

The overall pipeline consists of four stages. Firstly, we use existing pre-trained models or toolkits to extract the visual and audio features corresponding to each video frame in the videos. Secondly, each visual or audio feature sequence is input to the Affine Module to get features with the same dimension. Thirdly, these features are concat and then input to the Transformer Encoder we constructed to model the temporal features. Finally, the output of the encoder is input to the Output Layer to get the corresponding output. Figure \ref{fig:model} shows the overall framework of our proposed method.

The main notations used in this paper are listed as follows. Give a video, we can extract the visual features or audio features corresponding to all its video frames separately. We denote all the features by ${f_1,f_2,...,f_n}$, where $n$ is the number of features. For each of the four challenges, we denote their labels by $y^{va}, y^{expr}, y^{au}, y^{eri}$.

\subsection{Affine Module}
In our experiments, the inputs are one or several visual features or audio features, yet their dimensions are often different, even by a large margin. The dimensions of the features are shown in table \ref{tab:feat_dim}. We can see that the EAC feature have 2048 dimensions while FAU has only 17 dimensions. We think that too large a dimensional difference could diminish the effect of useful features. For this purpose, we design the Affine Module. For the former three challenges, we use Linear layers to affine the features of different dimension to the same dimension. Besides, following the setup of the classical Transformer\cite{AshishVaswani2017AttentionIA}, we add Position Encoding(PE) to each feature sequence for conveying its contextual temporal information. It can be formulated as follows:

\begin{equation}
  \hat{f}_i = (W_Af_i + b_A) + PE
  \label{eq:affine}
\end{equation}
where $W_A$ and $b_A$ are learnable parameters.

For the fourth challenge, we use 1D temporal convolution network to capture temporal information for each features. For the outputs, we also add the positional encoding.

\begin{table}
  \centering
  \begin{tabular}{l c c}
    \hline
    Feature & Modality & Dimension \\
    \hline
    IS09     & A & 384  \\
    CNN14    & A & 2048 \\
    VGGish   & A & 128  \\
    eGeMAPS  & A & 88   \\
    DeepSpectrum & A & 1024 \\
    EAC      & V & 2048 \\
    FAU      & V & 17   \\
    ResNet18 & V & 512  \\
    POSTER   & V & 768  \\
    POSTER2  & V & 768  \\
    \hline
  \end{tabular}
  \caption{The dimensions of features.}
  \label{tab:feat_dim}
\end{table}

\subsection{Transformer Encoder}
We model the temporal feature using the classical Transformer Encoder\cite{AshishVaswani2017AttentionIA}. As shown in Equation \ref{eq:concat} and Equation \ref{eq:transformer}, we first concat the outputs of the affine module and next input them to the Transformer Encoder for former three challenge. As shown in Equation \ref{eq:temma}, we use TEMMA model to obtain the temporal feature for the fourth challenge. It can be formulated as follows:
\begin{equation}
    \hat{f} = concat(\hat{f}_1, \hat{f}_2,...,\hat{f}_n)
    \label{eq:concat}
\end{equation}
\begin{equation}
    t = TransformerEncoder(\hat{f})
    \label{eq:transformer}
\end{equation}
\begin{equation}
    t = TEMMA(\hat{f})
    \label{eq:temma}
\end{equation}
where $t$ is the temporal feature.

\subsection{Output Layer}
After getting the temporal feature by Transformer Encoder, we input the feature $t$ to the Output Layer. The Output Layer consists of fully-connected layer, which can be formulated as follows:
\begin{equation}
    \hat{y} = W t + b
    \label{eq:output}
\end{equation}
which $W$ and $b$ are learnabel parameters. $\hat{y}$ is the predicton. 

\subsection{Loss Function}
Since each task is different, we apply different loss function. Next, we introduce them separately.

\subsubsection{Valence-Arousal(VA) Estimation Challenge}
We utilize the Mean Squared Error (MSE) as the loss function for the VA challenge, which can be formulated as:
\begin{equation}
    L(y,\hat{y}) = \frac{1}{N}\sum_{i=1}^N (y_i-\hat{y}_i)^2
    \label{eq:mse_va}
\end{equation}
which $y_i$ and $\hat{y}_i$ is the label and prediction of valence or arousal, $N$ is the number of frames in a batch.

\subsubsection{Expression (Expr) Classification Challenge}
We utilize the Cross Entropy (CE) as the loss function for the Expr challenge, which can be formulated as:
\begin{equation}
    L(y,\hat{y}) = -\sum_{i=1}^N \sum_{j=1}^C y_{ij} \log{\hat{y}_{ij}}
    \label{eq:crossentropy}
\end{equation}
which $y_{ij}$ and $\hat{y}_{ij}$ is the label and prediction of expression, $N$ is the number of frames in a batch and $C=8$ which denotes the number of expressions.

\subsubsection{Action Unit (AU) Detection Challenge}
We utilize the weighted asymmetric loss \cite{luo2022learning} as the loss function for the AU challenge, which can be formulated as:
\begin{equation}
    L(y,\hat{y}) = -\frac{1}{N} \sum_{i=1}^N w_i[y_i\log{\hat{y}_i}+(1-y_i)\hat{y}_i\log{(1-\hat{y}_i)}]
    \label{eq:auloss}
\end{equation}
which $\hat{y}_i$, $y_i$ and $w_i$ are the prediction (occurrence probability), ground truth and weight of the $i^{th}$ AU. By the way, $w_i$ is defined by the occurrence rate of the $i^{th}$ AU in the whole training set.

\subsubsection{Emotional Reaction Intensity (ERI) Estimation Challenge}
We also utilize the Mean Squared Error (MSE) as the loss function for the ERI challenge, which can be formulated as:
\begin{equation}
    L(y,\hat{y}) = \frac{1}{NC}\sum_{i=1}^N \sum_{j=1}^C (y_{ij}-\hat{y}_{ij})^2
    \label{eq:mse_eri}
\end{equation}
which $y_{ij}$ and $\hat{y}_{ij}$ is the label and prediction of expression, $N$ is the number of frames in a batch and $C=7$ which denotes the number of emotional reactions.

\section{Experiments}
\subsection{Dataset}
The 5th ABAW competition includes four challenges: 1) Valence-Arousal (VA) Estimation , 2) Expression (Expr) Classification, 3) Action Unit (AU) Detection, and 4) Emotional Reaction Intensity (ERI) Estimation. All challenges will accept only uni-task solutions. The second challenge involves 548 videos, while the third challenge involves 547 videos. Both challenges are based on Aff-Wild2, which is an audiovisual dataset containing approximately 2.7 million frames in total. For the first challenge an augmented version of the Aff-Wild2 database is used. This database consists of 594 videos of around 3M frames of 584 subjects annotated in terms of valence and arousal. 

As for the fourth challenge, the Hume-Reaction dataset is used. The dataset is a multimodal collection of approximately 75 hours of video recordings capturing 2222 subjects from South Africa and the United States reacting to a variety of emotional video stimuli in their homes via webcam.

We use the RAF-DB and AffectNet datasets for pre-training visual feature extractors. The RAF-DB is a large-scale database which consists of approximately 30,000 facial images from thousands of individuals. Each image has been annotated independently about 40 times and then filtered using the EM algorithm to remove unreliable annotations. AffectNet dataset is a large facial expression recognition dataset containing over one million facial images collected from the internet. About half of the images have been manually annotated for seven discrete facial expressions (neutral, happy, sad, angry, fearful, surprised, and disgusted) and the intensity of valence and arousal present in the facial expression.

\subsection{Experiment Setup}
For the former three challenge, due to the limitation of GPU memory, we segment each video with segment length set to 256. The batchsize size is 128, the output dimension of the affine module is 512 or 256, the number of encoder layers is 4, and the number of attention headers is 4. The feedforward and hidden layer dimensions are determined by the input dimension.

For the forth challenge, the number of convolutional layers is 5 and the kernel size is 3 in the input process block. The encoder blocks in the Multimodal encoder module is 4 and the number of heads in the multi-head attention layer is 4. For the inference module, the number of nodes in the last fully connected layer is 256 and the dropout is 0.2.

All the experiments are implemented with Pytorch. We adopt the Adam optimizer with the initial learning rate of 0.0001.

\subsection{Experimental Results}
\subsubsection{Valence-Arousal(VA) Estimation Challenge}
For the Valence-Arousal Estimation Challenge, table \ref{tab:va} shows the results of using single feature or using multiple features at the same time on the validation set.
\begin{table}
    \resizebox{\linewidth}{!}{
    \centering
    \begin{tabular}{l|c|c|c}
    \hline
    Features & Valence & Arousal & Mean \\
    \hline
    Baseline & 0.24 & 0.20 & 0.22 \\
    EAC & 0.4479 & 0.5878 & 0.5179 \\
    POSTER & 0.3920 & 0.6317 & 0.5119 \\
    ResNet18 & 0.4762 & 0.5671 & 0.5217 \\
    POSTER2 & 0.5374 & 0.6297 & 0.5836 \\
    ResNet18+VGGish & 0.4742 & 0.6220 & 0.5481 \\
    ResNet18+POSTER2 & 0.5515 & 0.6429 & 0.5972 \\
    ResNet18+POSTER2+FAU & 0.4868 & 0.6301 & 0.5585 \\
    POSTER2+POSTER+VGGish & 0.5003 & \textbf{0.6946} & 0.5975 \\
    EAC+ResNet18+POSTER2+VGGish & \textbf{0.5542} & 0.6590 & \textbf{0.6066} \\
    \hline
    \end{tabular}
    }
    \caption{The results on the validation set of Valence-Arousal Estimation Challenge.}
    \label{tab:va}
\end{table}

As can be seen in table \ref{tab:va}, for single features, POSTER2 performs best, and it has the best results in Valence, Arousal and Mean. It is obvious that for multiple features, the best combination of each value contains the feature POSTER2, which is a good indication of the effectiveness of the feature. In addition to this, we also find that the audio feature VGGish is useful for performance improvement.

\subsubsection{Expression (Expr) Classification Challenge}
For the Expression Classification Challenge, table \ref{tab:expr} shows the results of using single feature or using multiple features at the same time on the validation set.

For the Expression Classification Challenge, we can see that most of the features perform somewhat poorly in the classification task, both single and multiple features. The exception is the feature POSTER2, which extracts the expression information in the video quite well and achieves the best results with the F1 Score metric.

\begin{table}[htb]
    \centering
    \begin{tabular}{l|c}
    \hline
    Features & F1 \\
    \hline
    Baseline & 0.23\\
    EAC & 0.3188 \\
    POSTER & 0.3215 \\
    ResNet18 & 0.3176 \\
    POSTER2 & \textbf{0.4055} \\
    ResNet18+FAU & 0.3287 \\
    POSTER2+POSTER & 0.3630 \\
    ResNet18+POSTER2 & 0.3957 \\
    ResNet18+POSTER2+VGGish & 0.3580 \\
    EAC+ResNet18+POSTER2+VGGish & 0.3306 \\
    \hline
    \end{tabular}
    \caption{The results on the validation set of Expression Classification Challenge.}
    \label{tab:expr}
\end{table}

\subsubsection{Action Unit (AU) Detection Challenge}
For the Action Unit Detection Challenge, table \ref{tab:au} shows the results of using single feature or using multiple features at the same time on the validation set.
\begin{table}[htb]
    \centering
    \begin{tabular}{l|c}
    \hline
    Features & F1 \\
    \hline
    Baseline & 0.39 \\
    EAC & 0.4881 \\
    POSTER & 0.5046 \\
    ResNet18 & 0.5114 \\
    POSTER2 & 0.5181 \\
    ResNet18+FAU & 0.5079 \\
    POSTER2+FAU & \textbf{0.5296} \\
    POSTER2+POSTER & 0.5112 \\
    ResNet18+POSTER2 & 0.5247 \\
    ResNet18+POSTER2+VGGish & 0.5014 \\
    EAC+ResNet18+POSTER2+VGGish & 0.4949 \\
    \hline
    \end{tabular}
    \caption{The results on the validation set of Action Unit Detection Challenge.}
    \label{tab:au}
\end{table}

For the Action Unit Detection Challenge, we can see that the performance of all features in the detection task is relatively close for both single and multi-features. It is obvious that FAU features can have some effect, while audio features are not effective. Our best result come from POSTER2 and FAU.

\subsubsection{Emotional Reaction Intensity (ERI) Estimation Challenge}
For the Emotional Reaction Intensity Estimation Challenge, table \ref{tab:eri} shows the results of using single feature or using multiple features at the same time on the validation set.
\begin{table}[htb]
    \centering
    \begin{tabular}{l|c}
    \hline
    Features & PCC \\
    \hline
    Baseline & 0.2488 \\
    CNN14 & 0.1582 \\
    ResNet18 & 0.3893 \\
    eGeMAPS & 0.0733 \\
    DeepSpectrum & 0.1835 \\
    ResNet18+CNN14 & 0.3839 \\
    ResNet18+eGeMAPS & 0.3809 \\
    ResNet18+DeepSpectrum & 0.3968 \\
    \hline
    \end{tabular}
    \caption{The results on the validation set of Emotional Reaction Intensity Estimation Challenge.}
    \label{tab:eri}
\end{table}

The combination of ResNet18 features with low-level audio features such as eGeMAPS and CNN14 gives poorer results than the performance of ResNet18 features, suggesting that they are not effective. The combination of ResNet18 and DeepSpectrum yields better results. This shows that these two features can complement each other to provide more comprehensive expression information.

\section{Conclusion}
In this paper, we present our solutions for the 5th Workshop and Competition on Affective Behavior Analysis in-the-wild (ABAW), which includes four sub-challenges of Valence-Arousal (VA) Estimation Challenge, Expression (Expr) Classification Challenge, Action Unit (AU) Detection Challenge and Emotional Reaction Intensity (ERI) Estimation Challenge. We extract powerful audio and visual features and designed an Affine Module to mitigate the impact of different features. Extensive experiments demonstrate that our method significantly outperforms the baseline and achieves excellent results in the four sub-challenges.

{\small
	
	\bibliography{aaai}
}

\end{document}